\documentclass[final]{OAGM}
\OAGMarXiv{1304.1876}

\usepackage[utf8]{inputenc}
\usepackage{amsmath}

\usepackage{todonotes}

\usepackage{graphicx}
\usepackage{epstopdf}
\usepackage{gensymb}
\usepackage{acronym}

\title{Digit Recognition in Handwritten Weather Records}

\author{Manuel Keglevic and Robert Sablatnig\\
  Computer Vision Lab, Vienna University of Technology, Austria}

\acrodef{ocr}[\textsc{OCR}]{Optical Character Recogntion}
\acrodef{cc}[\textsc{cc}]{Connected Component}
\acrodef{svm}[\textsc{SVM}]{Support Vector Machine}
\acrodef{rbf}[\textsc{rbf}]{Radial Basis Function}
\acrodef{roi}[\textsc{roi}]{Region of Interest}
\acrodef{pca}[\textsc{PCA}]{Principle Component Analysis}
\acrodef{dscc}[\textsc{DSCC}]{Directional Single-Connected Chain}
\acrodef{cnn}		[\textsc{CNN}]		{Convolutional Neural Network}
\acrodef{knn}		[\textsc{k-NN}]		{k-Nearest Neighbor}	
\acrodef{nn}		[\textsc{NN}]		{Neural Network}
\acrodef{mqdf}		[\textsc{MQDF}]		{Modified Quadratic Discriminant Function}

\begin{document}
\maketitle

\begin{abstract}

This paper addresses the automatic recognition of handwritten temperature values in weather records. The localization of table cells is based on line detection using projection profiles. Further, a stroke-preserving line removal method which is based on gradient images is proposed. The presented digit recognition utilizes features which are extracted using a set of filters and a Support Vector Machine classifier. It was evaluated on the MNIST and the USPS dataset and our own database with about 17,000 RGB digit images. An accuracy of 99.36\% per digit is achieved for the entire system using a set of 84 weather records.

\end{abstract}

\section{Introduction}

In addition to handwritten documents stored in historic archives there are processes, like manually filling out forms, which still depend on pen and paper. However, accessing the information stored in those documents requires time and manpower~\cite{Richarz:2012}. By digitizing those documents, the advantages of digitally stored information, such as ease of access, can be exploited for handwritten documents~\cite{Richarz:2012}. 

\begin{figure}[bt]
	\centering
	\includegraphics[width=\linewidth]{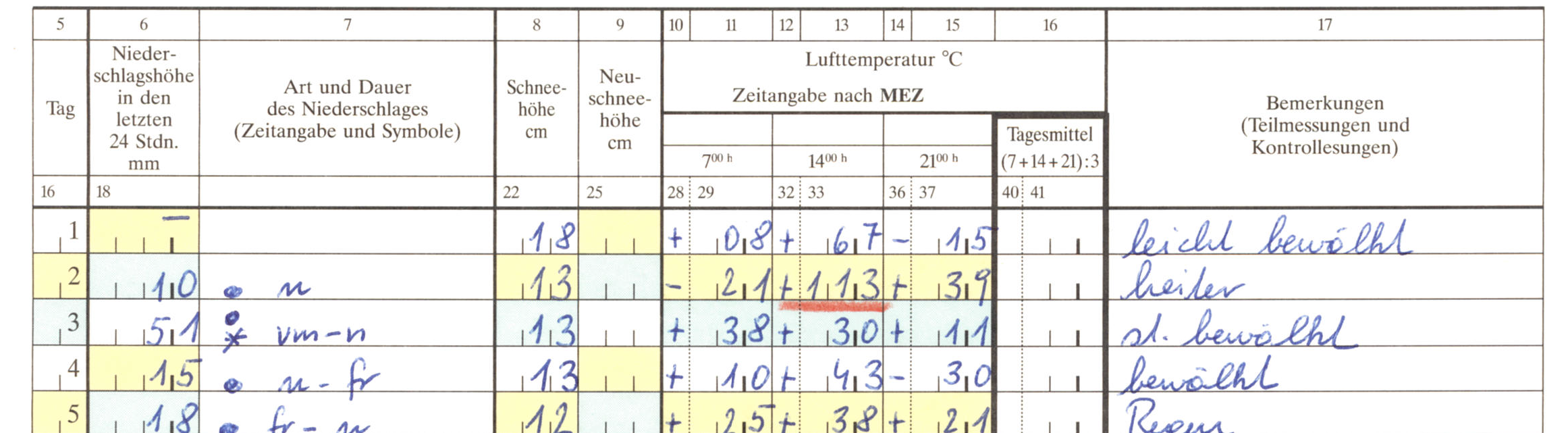}
	\caption{Weather record from February of the year 2000 from a measurement station in Lower Austria. The temperature values are located in the column "Lufttemperatur \degree C".}
	\label{fig:weather-record}
\end{figure}

The documents covered in this paper, i.e. weather records, consist of known printed forms with handwritten information. Even though the structure of the form is known beforehand, due to prior handling and the scanning process, global and local deformations are introduced. Further, the structure of forms suggests certain areas which are applicable for handwriting, yet writers are not bound to this constraint. Additionally, handwriting varies significantly due to different writing styles. Even the handwriting of a single writer exhibits variations. In Figure~\ref{fig:weather-record} an example of a weather record is shown. The origins of the weather records regarded in this paper are several measurement stations located in Austria (Lower Austria). 

The localization of the numerical data is conducted by first reconstructing the tabular structure of the form. Using vertical and horizontal projection profiles, the rough positions of the lines building up the table are found. Errors in the reconstruction are detected using so-called \textit{Form Properties} which store a-propri information of the form. Finally, the extraction of the digits and signs is done using a binarization based on the Savakis filter~\cite{Savakis:1998} and subsequent \ac{cc} analysis.

The handwritten digit recognition is based on the work of Labusch et al.~\cite{Labusch:2008}. Using \ac{pca} the features of the normalized characters are extracted. As a classifier a multiclass \ac{rbf} \ac{svm} is used. 

The paper is structured as follows: in the following section digit recognition and document analysis for digit recognition is presented. In Section~\ref{sec:da-dr} the localization of the numerical data is explained in detail. Subsequently, the digit recognition approach is shown in Section~\ref{sec:dr}. Finally, in Section~\ref{sec:eval} an evaluation of the presented method is conducted and final remarks are given in Section~\ref{sec:con}.

\section{Related Work}

Earlier systems for recognizing handwritten digits in documents include for instance US addresses~\cite{Srihari:1993} and census forms~\cite{Garris:1995}. More recently, Bulacu et al.~\cite{Bulacu:2009} presented a system for recognizing handwritten digits in historic documents from the archive of the Cabinet of the Dutch Queen. Further, Richarz et al.~\cite{Richarz:2012} proposed a semi-supervised method for the transcription of historic weather documents. 

For digit recognition Liu et al.~\cite{Liu:2005} presented an approach using local stroke directions of the handwritten digits as features. These features are extracted from gradient directions. 
Teow et al.~\cite{Teow:2002} proposed a feature extraction approach which is inspired by the biologic vision system. They use 16 filters designed to detect edges and end-stops of various orientations. These filters simulate the behaviour of receptive fields in the visual system. 
Another vision based approach is proposed by Lambusch et al.~\cite{Labusch:2008}. They proposed a feature extraction method based on learned sparse representations. 
Keysers et al.~\cite{Keysers:2007} proposed an image matching approach for digit recognition. The main idea is to map the pixels of a test image onto the pixels of a reference image. The quality of the mapping is determined by a distance function which is used in combination with a k-NN classifier to predict the class label of the test image. 

As shown by Lecun et al.~\cite{Lecun:1998} it is possible to achieve state-of-the-art recognition accuracies without using a hand-crafted feature extractor but instead incorporate the feature extraction process in the (trainable) classifier. They proposed a specialized \ac{nn}, the so-called \ac{cnn}, with alternating convolution and subsampling layers.
Ciresan et al. further improved this architecture by proposing so-called Deep \ac{nn} and by combination of multiple Deep \ac{nn}~\cite{Ciresan:2012}.

\section{Document Analysis for Digit Recognition}
\label{sec:da-dr}
In order to extract the fields containing the numerical data, i.e. the signs and digits, they have to be located beforehand. The documents in scope of this paper, i.e. the weather records, are forms which have been filled out manually. Therefore, an underlying tabular structure is present within the documents. In this section, first an approach using line detection for reconstructing the form is presented. Second, a method for stroke preserving removal of form lines is shown. Third, the segmentation of the digits and signs is depicted.

\subsection{Table Reconstruction}
The first step for reconstructing the table structure is finding vertical and horizontal lines in the image. Starting from a rotation corrected grayscale image~\cite{Diem:2012} vertical and horizontal projection profiles are used to find the vertical and horizontal lines, respectively. 

\begin{figure}[tb]
	\centering
	\includegraphics[width=\linewidth]{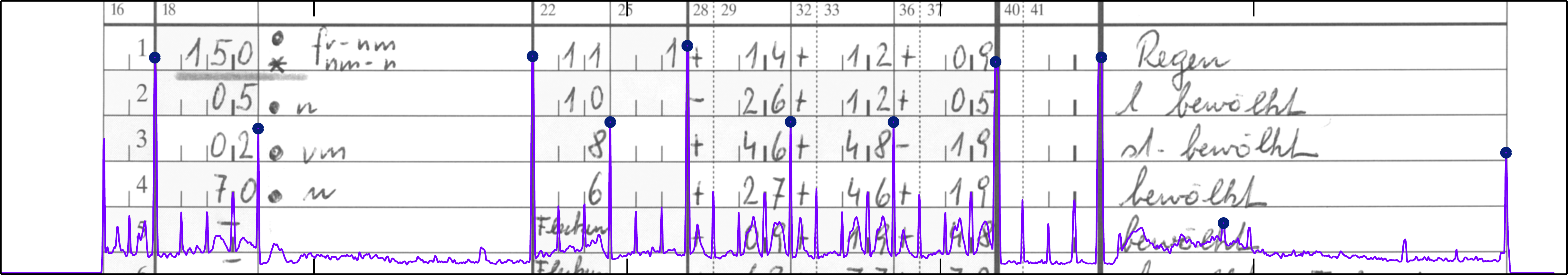}
	\caption{Image showing vertical projection profile with the detected peaks and the part of image in background.}
	\label{fig:vertical-projection}
\end{figure}

As shown in Figure~\ref{fig:vertical-projection} the peaks in the vertical projection correspond to the vertical lines. By introducing a minimum horizontal distance the peaks are searched in local areas. The blue circles in Figure~\ref{fig:vertical-projection} depict the peaks detected which are used for further processing. 

To cope with errors in the line detection process, a-priori information about the form, so-called \textit{Form Properties}, is used to correct the detected lines. For instance, horizontal lines are added if there are gaps contradicting the \textit{Form Properties}. 

\subsection{Line Removal}
To segment the single digits and signs it is necessary to remove the form borders. However, due to cursively written digits and signs, part of the symbols may cross the borders. To preserve the characteristics of the digits, a line removal preserving handwritten strokes is applied.

This is done using a Wiener filter on vertical and horizontal gradient images. The Wiener filter is a reconstruction filter. Starting from a filtered - or degraded - image ${G}(u,v)$, which is filtered with a known filter function $H(u,v)$ (in this case a vertical or horizontal Sobel filter), the task is to reproduce the original unfiltered image $F(u,v)$. 

To cope with noise, the Wiener filter extends the inverse filtering approach by modelling the noise and the image as two random variables. 
The Wiener filter minimizes the mean square error between the reconstructed image $\hat{F}(u,v)$ and the original image $F(u,v)$. This minimum is given in the frequency domain as 

\begin{equation} \label{eq:wiener}
	\hat{F}(u,v) = \left[\frac{1}{H(u,v)}\frac{ {\lvert H(u,v) \rvert}^2}{{\lvert H(u,v) \rvert}^2 + K}\right] G(u,v)
\end{equation}

where ${\lvert H(u,v) \rvert}^2$ is defined as the product of the filter function with its complex conjugate. The parameter $K$ represents the signal-to-noise ratio, i.e. the power spectrum of the noise ${\lvert N(u,v) \rvert}^2$ divided by the power spectrum of the original image ${\lvert F(u,v) \rvert}^2$. The term in the brackets in Equation \ref{eq:wiener}
 is called the Wiener filter. 

\begin{figure}[h]
	\label{fig:border-removal}
	\centering
	\begin{tabular}{c c c c c}
	\includegraphics[width=0.16\linewidth]{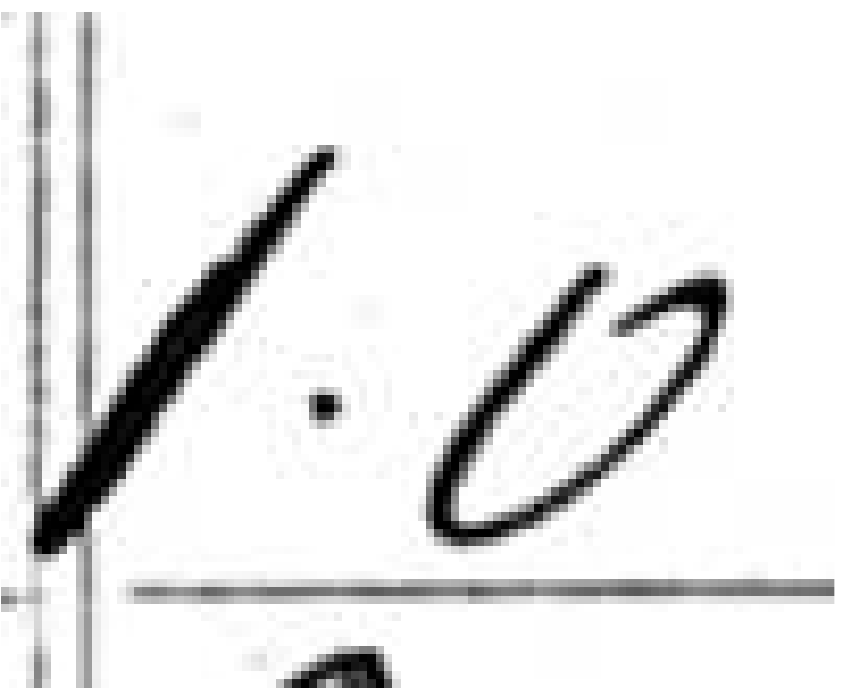} &
	\includegraphics[width=0.16\linewidth]{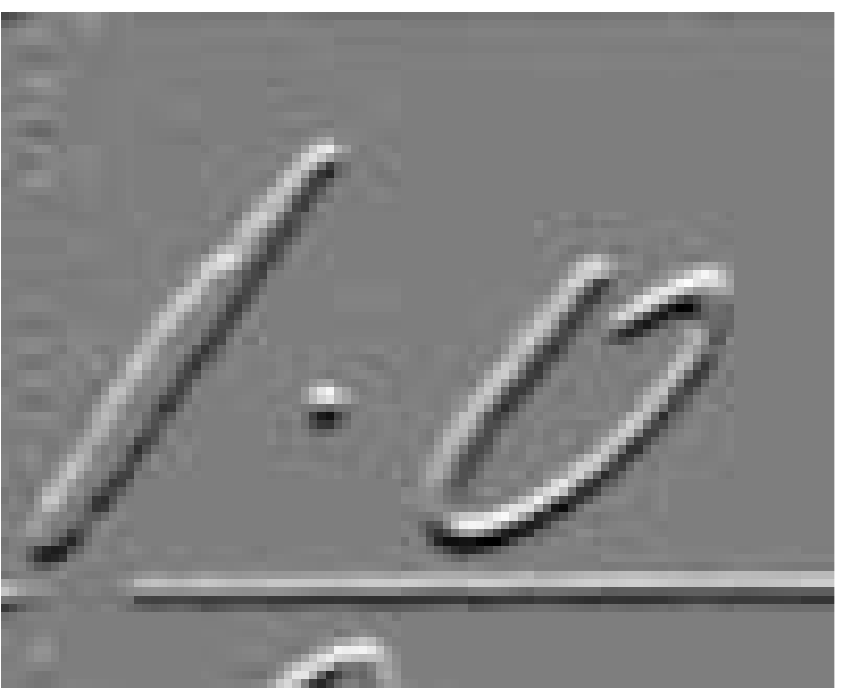} &
	\includegraphics[width=0.16\linewidth]{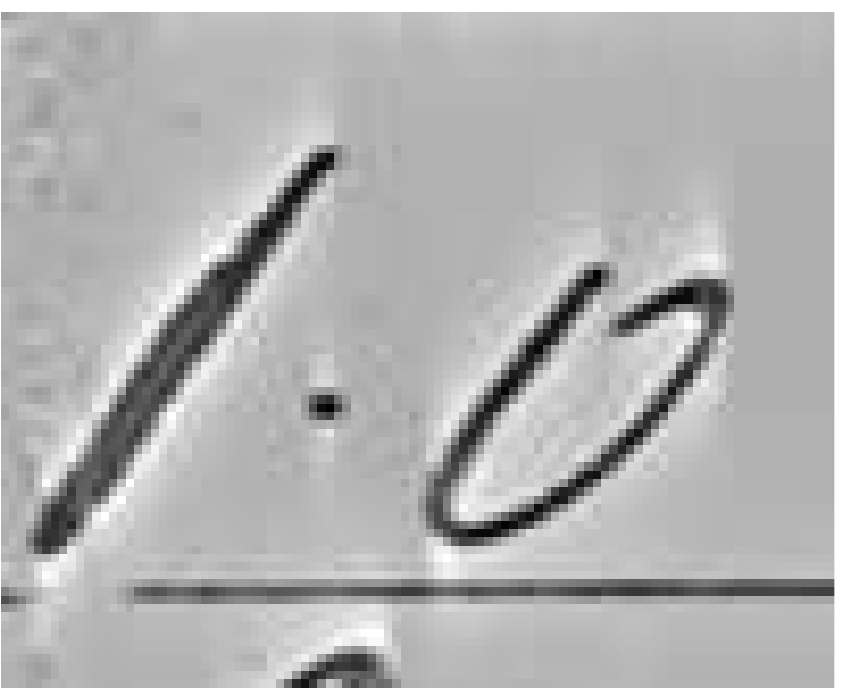} &
	\includegraphics[width=0.16\linewidth]{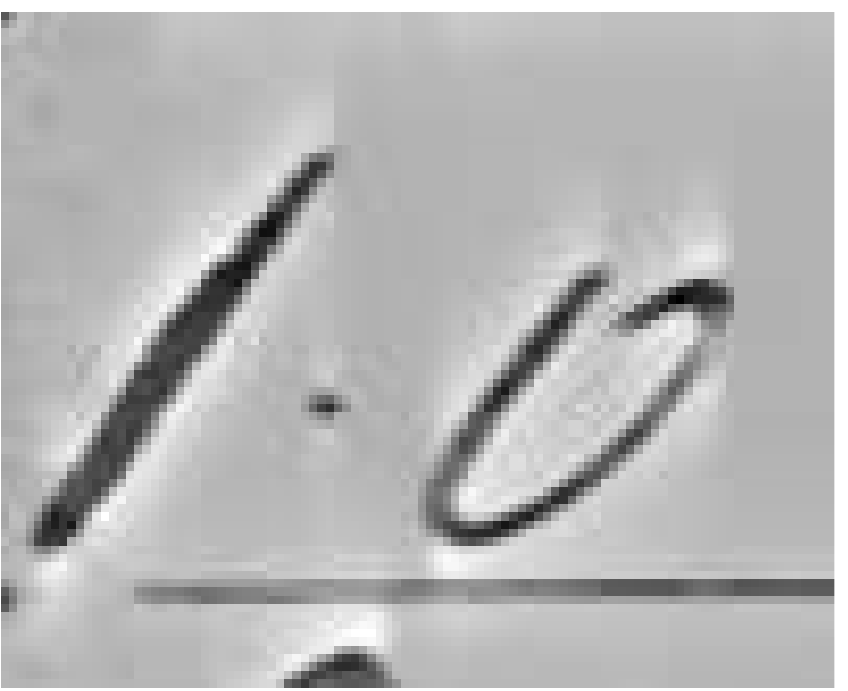} & 
	\includegraphics[width=0.16\linewidth]{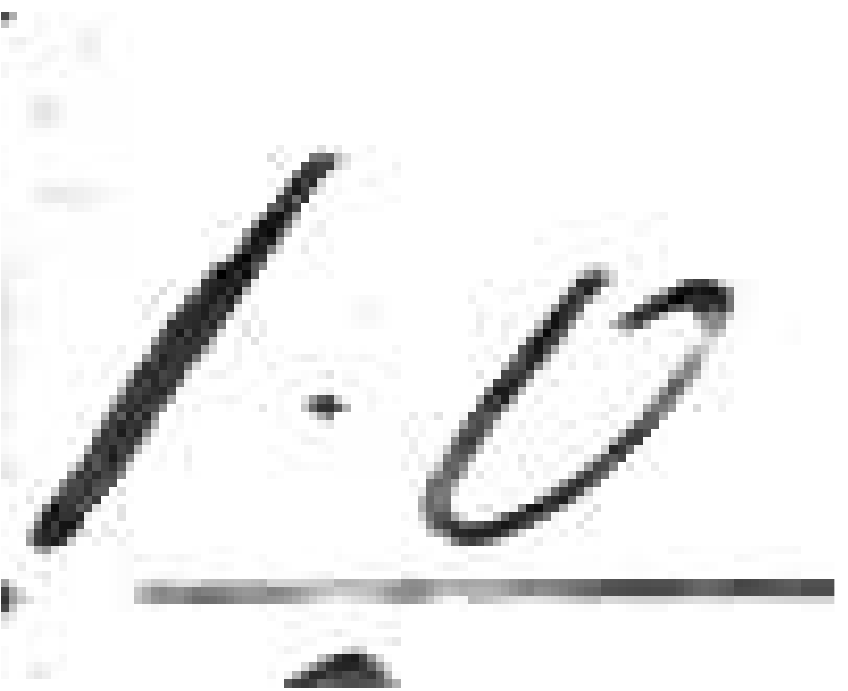} \\
	(a) & (b) & (c) & (d) & (e)
	\end{tabular}
	\caption{The five states of border removal using the Wiener filter~(a-e). The original image (a), gradient image (in this case vertical) (b), gradient after deconvolution (c), median filtered deconvolution (d) and resulting image after scaling the intensity values~(e). }
	\label{fig:edge_removal}
\end{figure}

The parameter $K$ can be used to limit the reconstruction effect of the Wiener filter. 
 Additionally, by computing the gradient the constant term of the image is lost which is advantageous in case of border removal. 
 In Figure~\ref{fig:edge_removal} the whole process is visualized. In order to remove the vertical lines in the original image (a), first the vertical gradient is computed. As shown in (b) this removes the vertical lines. To reconstruct the digits, the Wiener filter is used. In (c) the inner area of the digit `1' is reconstructed while the vertical lines remain removed. After that, median filtering (see (d)) and intensity scaling is used to remove artifacts and enhance the result (see (e)).  

However, only the regions around the vertical and horizontal lines have to be altered. This means, in the final image, these are the only regions which are replaced while the rest of the image stays the same. This way, regions not affected by horizontal or vertical lines retain the initial image quality. 

The rough line regions are defined by the reconstructed form and, additionally, an offset is used to extend the search area to cope with local variations. The exact location of the lines is subsequently found by searching for peaks in the gradient images. This is carried out locally for each column and row in the \ac{roi}, respectively. 

\subsection{Segmentation}

In the next step the background is removed using a threshold image. The threshold image is computed using an adapted version of the Savakis filter~\cite{Savakis:1998}. In distinct 15x15 blocks the image pixels are grouped into foreground pixels $v_i$ and background pixels $w_j$ using a precomputed threshold (Otsu). The local threshold $t$ is then defined as the boundary separating the mean of the intensity values of the foreground pixels $\overline{v}$ and the mean of the intensity values of the background pixels $\overline{w}$:

\begin{equation}
	t = \frac{\overline{v} + \overline{w}}{2},\; \text{with}\; \overline{v} = \frac{1}{I} \sum_{i=1}^{I} v_i \; \text{and} \; \overline{w} = \frac{1}{J} \sum_{j=1}^{J} v_j
\end{equation}

\begin{figure}[htb]
	\centering
	\begin{tabular}{c c c c}
	\includegraphics[width=0.15\linewidth]{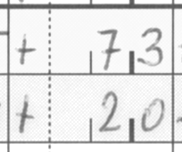} &
	\includegraphics[width=0.15\linewidth]{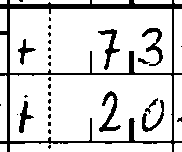} &
	\includegraphics[width=0.15\linewidth]{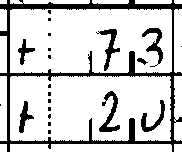} &
	\includegraphics[width=0.15\linewidth]{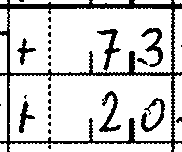} \\
	(a) & (b) & (c) & (d) \\
	\end{tabular}
	\caption{Comparison of different binarization techniques on an example (a). The method proposed (b) and approaches by Su et al. from the years 2010~\cite{Su:2010} (c) and 2011~\cite{Su:2011} (d).}
	\label{fig:binarization}
\end{figure}

In Figure~\ref{fig:binarization} this approach is compared to two state-of-the-art binarization approaches. Due to the high image quality of the weather records, this approach shows less missing character strokes and clearer frame boarders. The threshold image is then used to remove the background pixels. Additionally, a Gaussian filter is applied to enhance the result.

The digits and signs are then extracted by computing \acfp{cc} and assigning them to the nearest cell centers. To remove noise, \acp{cc} where the number of pixels is under a certain threshold are removed. The last step before the character recognition is a size normalization. 

\section{Handwritten Digit Recognition}
\label{sec:dr}

The digit recognition algorithm is based on the work of Labusch et al.~\cite{Labusch:2008}. In this section, first the calculation of the basis functions is explained. Then, the feature extraction and classification are outlined.

\subsection{Basis Functions}

The basis functions are learned using 100,000 image patches. As proposed in~\cite{Labusch:2008} the size of the image patches is set to 13x13.  A patch $P(x,y)$ is extracted by placing it at a random position in a training image. The training images used for this extraction are evenly distributed among the whole training set. As stated by Labusch et al.~\cite{Labusch:2008} preprocessing is necessary for the PCA. This is done by converting the extracted patches into so-called \textit{centered vectors}~\cite{Labusch:2008}. First, the mean pixel value of the patch is subtracted: 

\begin{equation}
	S(x,y) = P(x,y) - \overline{P}
\end{equation}

Next, $\overline{S(x,y)}$ the mean value over all $S(x,y)$ is computed for each pixel. Finally, the centered vectors are obtained by subtracting $\overline{S(x,y)}$ from all $S(x,y)$

\begin{equation}
	S_{p}(x,y) = S(x,y) - \overline{S(x,y)}
\end{equation}

After extracting the patches, a PCA is used to learn the basis functions from these samples, i.e. the parameter of the underlying model. This is done by creating the covariance matrix and computing the eigenvectors and the corresponding eigenvalues.

\subsection{Feature Extraction and Classification}
The coefficient images are computed by subsequently convoluting the image with the computed eigenvectors. As a result, for each of the 13x13 basis functions, a coefficient image $F_i$ is produced. To achieve local shift invariance the extrema in 9 distinct blocks in each of the $F_i$ are used as feature vector.

As proposed by Labusch et al.~\cite{Labusch:2008}, an \ac{rbf} is chosen as the kernel for the \ac{svm}. The one-versus-one approach is used for handling multiple classes. This is important for digit recognition due to the fact that  there are at least ten classes and \acp{svm} without this extension only allow binary classification.

\section{Experiments}
\label{sec:eval}

In this section first the digit recognition module is evaluated using the MNIST, USPS and our own digit database. Subsequently, the whole process starting with scanned images to classified digits is evaluated on a set of weather records.

\subsection{Digit Recognition}

For the evaluation of the digit recognition module three different datasets are used. First, the MNIST\footnote{http://yann.lecun.com/exdb/mnist/} database which consists of 60,000 training images and 10,000 test images.
Second, the USPS dataset consists of 7,291 training samples and 2,007 test samples. 
Third, digit images of our own database were used as comparison. In contrast to the previous datasets, the images are available in RGB and no normalization procedure was applied beforehand. The digits were written by approximately 120 different writers. The 17,322 digit images in this database are randomly split into a trainingset containing 13,322 samples and a testset with 4,000 samples.

The digit recognition module achieves an accuracy of 99.24\% on the MNIST database. This coincides with the results reported by Lambusch et al.~\cite{Labusch:2008} on whom the method is based on. For the more challenging USPS database an accuracy of 97.16\% is achieved. Using the normalization proposed by the MNIST dataset an accuracy of 96,45\% is achieved on our own database. Using the probability values of the \ac{svm} to compute a second guess, the accuracy is increased to 99,05\%. This leads to the assumption that using these probabilities in combination with semantic information, such as mean and standard variation for the temperature values at a specific time and day, can lead to more accurate results. 

To investigate how the size difference of the training- and testset influences the results, an evaluation was conducted using swapped sets. On both the MNIST and our own database this leads to a decrease of the accuracy to 98.20\% and 94.75\%, respectively.

\subsection{Weather records}

The weather records used for evaluation were recorded in the years 1994 to 2000 at a measuring station in the province Lower Austria. Each of the 84 documents contains the temperature values for one month with three measurements per day. In Figure~\ref{fig:weather-record} an example is shown. It has to be noted, that all records were filled out by the same writer and that for all documents a manually created ground-truth is available. However, it contains just the consolidated numbers though no information about the digits or signs itself. For instance, the plus sign is often omitted by writers (more than half of the time in these records) and leading zeros can be left out. To cope with this problem empty fields are treated either as an additional class or plus signs in case of digit and sign recognition, respectively. The accuracy of the digit recognition is computed using:

\begin{equation}
	\mbox{accuracy} = \frac{\mbox{number of correctly classified cells}}{\mbox{number of cells containing \acp{cc}}}
\end{equation} 

In Table~\ref{tab:acc} the accuracies are depicted for two different splits of the documents. Independent of the size difference between the training and testset the accuracy is above 99\%. Unfortunately, for the sign recognition there is no ground-truth stating the amount of plus signs present in the documents. However, in the first case depicted in Table~\ref{tab:acc} only three instances of falsely classified signs are found for the whole year 1994. Assuming one sign per number, this corresponds to an error rate of 0.3\%. 

\begin{table}[htb]
\centering
\begin{tabular}{ | c | c | c | c |}
\hline
	Trained Years & Tested Years & Tested Digit Cells & Digit Accuracy \\ \hline
	1995-2000 & 1994 & 208 & 99.10\% \\
	1997-2000 & 1994-1997 & 8,065 & 99.36\% \\ \hline
\end{tabular}
\caption{Digit cell recognition accuracies achieved using the weather records from the years denoted in the first two columns.}
\label{tab:acc}
\end{table}

\begin{figure}[htb]
   \centering
   \includegraphics[width=0.7\linewidth]{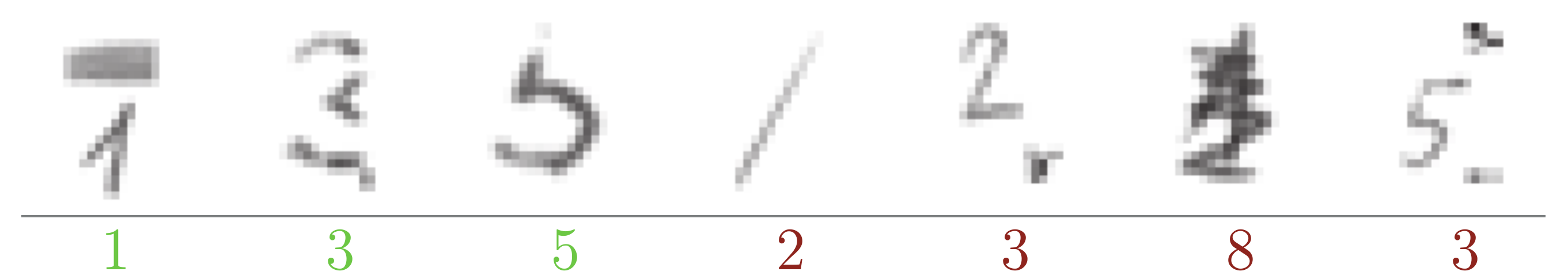}
   \caption{Examples of correctly (green) and falsely (red) classified normalized digit cells. The numbers at the bottom depict the output of the classification.}
   \label{fig:digits}
\end{figure}

The main source of error is the segmentation process. First, artifacts not removed before the normalization stage may lead to degraded normalized images. This is for instance shown in the 5th and 7th example in Figure~\ref{fig:digits}. Second, parts of the digit may be missing if either the line removal algorithm is not able to preserve the handwritten strokes or in case of under-segmentation. However, the 2nd and 3rd digit in Figure~\ref{fig:digits} show that a correct classification is nevertheless possible in some cases. Other problems are underlinings of temperatures, corrections and meaningless strokes which are shown in the 1st, 4th and 6th example in Figure~\ref{fig:digits}, respectively.

\section{Conclusions}
\label{sec:con}

In this paper an approach for the automatic recognition of temperature values in weather records is introduced. The segmentation of the digits is achieved by reconstructing the tabular structure using line detection. Using a filterbank the features of the digits and signs are extracted and subsequently classified with an \ac{svm}. The evaluation conducted on 84 weather records showed an accuracy of over 99\% per digit. The digit recognition model is able to cope with missing digit strokes and underlinings. However, due to the normalization process the impact of artifacts can be amplified and additionally the system is not able to cope with meaningless strokes or corrections. 

For future work, the main performance enhancements are expected by making the segmentation process more robust to artifacts. 

\section*{Acknowledgments}

We would like to thank  Gerhard Kubu and Maximilian Heilig for providing the weather records. This work was supported by \textit{die Buben}. 

\bibliography{bibtex}
\end{document}